\documentclass[runningheads]{llncs}
\usepackage{graphicx}

\usepackage{tikz}
\usepackage{comment}
\usepackage{amsmath,amssymb} 
\usepackage{color}

\usepackage[accsupp]{axessibility}  

\usepackage{caption}
\usepackage{subcaption}

\usepackage{url}

\begin{document}

\pagestyle{headings}
\mainmatter
\def\ECCVSubNumber{1}  

\title{CenDerNet: Center and Curvature Representations for Render-and-Compare 6D Pose Estimation}

\titlerunning{CenDerNet}
%
\author{
    Peter De Roovere\inst{1,2}\and 
    Rembert Daems\inst{1,3,4}\and 
    Jonathan Croenen\inst{2}\and 
    Taoufik Bourgana\inst{5}\and 
    Joris de Hoog\inst{5}\and 
    Francis wyffels\inst{1}
}
\authorrunning{P. De Roovere et al.}
%
\institute{IDLab-AIRO -- Ghent University -- imec \and
    RoboJob\and
    dept. Electromechanical, Systems and Metal Engineering -- Ghent University \and
    EEDT-DC -- Flanders Make \and
    DecisionS -- Flanders Make}
\maketitle

\begin{abstract}
    We introduce CenDerNet, a framework for 6D pose estimation from multi-view images based on center and curvature representations.
    Finding precise poses for reflective, textureless objects is a key challenge for industrial robotics.
    Our approach consists of three stages:
    First, a fully convolutional neural network predicts center and curvature heatmaps for each view;
    Second, center heatmaps are used to detect object instances and find their 3D centers;
    Third, 6D object poses are estimated using 3D centers and curvature heatmaps.
    By jointly optimizing poses across views using a render-and-compare approach, our method naturally handles occlusions and object symmetries.
    We show that CenDerNet outperforms previous methods on two industry-relevant datasets: DIMO and T-LESS.
\end{abstract}

\begin{figure}[ht]
    \centering
    \includegraphics[width=\linewidth,keepaspectratio]{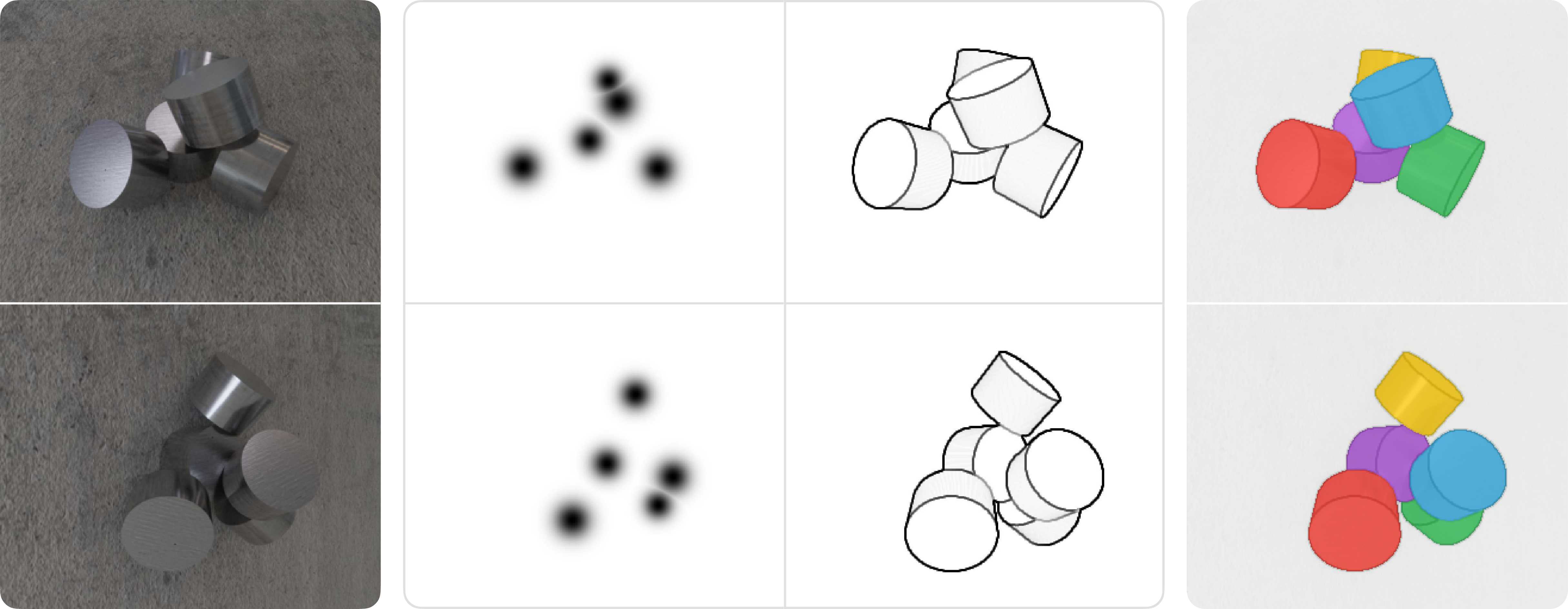}
    \caption{
        Multi-view input images (left) are converted to center and curvature heatmaps (center), used to estimate 6D object poses (right).
        \label{fig:main}
    }
\end{figure}

\section{Introduction}

\subsection{Context}

6D pose estimation is an essential aspect of industrial robotics.
Today's high-volume production lines are powered by robots reliably executing repetitive movements.
However, as the manufacturing industry shifts towards high-mix, low-volume production, there is a growing need for robots that can handle more variability \cite{doi/10.2873/34695}.
Estimating 6D poses for diverse sets of objects is crucial to that goal.

Manufacturing use cases present unique challenges.
Many industrial objects are reflective and textureless, with scratches or saw patterns affecting their appearance \cite{yang2021robi,deroovere2022dataset}.
Parts are often stacked in dense compositions, with many occlusions.
These densely stacked, reflective parts are problematic for existing depth sensors.
In addition, object shapes vary greatly, often exhibiting symmetries leading to ambiguous poses.
Many applications require sub-millimeter precision and the ability to integrate new, unseen parts swiftly.

This work presents a framework for 6D pose estimation targeted to these conditions.
Our approach predicts object poses for known textureless parts from RGB images with known camera intrinsics and extrinsics.
We use multi-view data as monocular images suffer from ambiguities in appearance and depth.
In practice, images from multiple viewpoints are easily collected using multi-camera or hand-in-eye setups.
This setup reflects many real-world industrial use-cases.

\subsection{Related work}

Recent progress in 6D pose estimation from RGB images was demonstrated in the 2020 BOP challenge~\cite{hodavn2020bop}.
Convolutional neural networks (CNNs) trained on large amounts of synthetic data are at the core of this success.
Many recent methods consist of three stages: (1) object detection, (2) pose estimation, and (3) refinement, using separate neural networks for each stage.
While most approaches operate on monocular RGB images, some focus on multi-view data.

\paragraph{Object detection}
In the first stage, CNN-based neural networks are used to detect object instances.
Although these detectors typically represent objects by 2D bounding boxes~\cite{fu2019retinamask,he2017mask,lin2017focal,ren2015faster}, 2D center points can be a simple and efficient alternative~\cite{zhou2019objects}.

\paragraph{Pose estimation}
In the second stage, the 6D pose of each detected object is predicted.
Classical methods based on local features~\cite{bay2006surf} or template matching~\cite{hinterstoisser2011multimodal} have been replaced by learning systems.
CNNs are used to detect local features~\cite{xiang2017posecnn,tekin2018real,tremblay2018deep,labbe2020cosypose} or find 2D-3D correspondences~\cite{peng2019pvnet,park2019pix2pose,zakharov2019dpod,wang2019normalized}.
Crucial aspects are the parameterization of 6D poses~\cite{zhou2019continuity} and how symmetries are handled~\cite{pitteri2019object}.

\paragraph{Refinement}
In the final stage, the estimated poses are iteratively refined by comparing object renders to the original image.
This comparison is not trivial, as real-world images are affected by lighting, texture, and background changes, not captured in the corresponding renders.
However, CNNs can be trained to perform this task~\cite{labbe2020cosypose,li2018deepim,zakharov2019dpod}.

\paragraph{Multi-view}
Approaches for multi-view 6D object pose estimation extend existing single-view methods.
First, poses hypotheses are generated from individual images. Next, these estimates are fused across views~\cite{labbe2020cosypose,li2018unified}.

\subsection{Contributions}

We present CenDerNet, a framework for 6D pose estimation from multi-view images based on center and curvature representations.
First, a convolutional neural network is trained to predict center and curvature heatmaps.
Second, center heatmaps are used to detect object instances and find their 3D centers.
These centers initialize and constrain the pose esimates.
Third, curvature heatmaps are used to optimize these poses further using a render-and-compare approach.


Our system is conceptually simple and easy to use.
Many existing methods consist of multiple stages, each with different training and tuning requirements.
Our framework is more straightforward.
We use a single, fully-convolutional neural network to convert RGB images to interpretable representations.
Next, we use classical optimization techniques that require little tuning.

Using a render-and-compare approach, we jointly estimate poses for all objects in a scene across all viewpoints.
As a result, our method naturally handles occlusions and object symmetries.
We provide a GPU implementation of our render-and-compare method that allows evaluating over 2,000 scene pose estimates per second.

We evaluate CenDerNet using DIMO and T-LESS, two challenging, industry-relevant datasets.
On DIMO, our method outperforms PVNet by a large margin.
On T-LESS, CenDerNet outperforms the 2020 ECCV results of CosyPose, the leading multi-view method.

\section{CenDerNet}

Our system consists of three stages:

\begin{enumerate}
    \item
          A convolutional neural network predicts center and curvature heatmaps for multi-view input images.
    \item
          The predicted center heatmaps are converted to 3D center points.
          These 3D centers initialize and constrain the set of predicted object poses.
    \item
          Object poses are optimized by comparing curvature renders to the predicted curvature heatmaps.
\end{enumerate}

\subsection{From images to center and curvature heatmaps}

\begin{figure}[htb!]
    \centering
    \includegraphics[width=\linewidth,keepaspectratio]{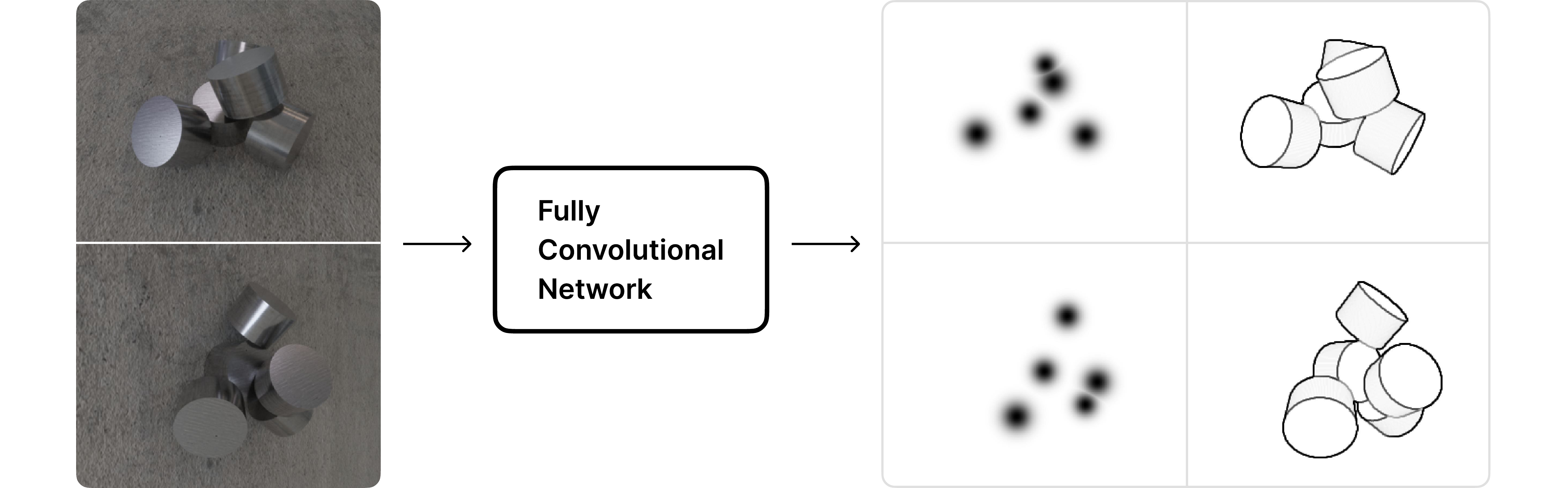}
    \caption{
        Step 1: multi-view RGB images are converted to center and curvature heatmaps by a single, fully convolutional network.
        \label{fig:step_1}
    }
\end{figure}

This step eliminates task-irrelevant variations by converting images into center and curvature representations.
RGB images can vary significantly due to lighting, background, and texture changes.
These changes, however, do not affect object poses.
This step eliminates these effects by transforming images into representations that simplify pose estimation.
As we want our system to apply to a wide range of objects, these representations should be category-independent.
We identify centers and curvatures as suitable representations with complementary properties.

\paragraph{Centers}
We use center heatmaps --- modeling the probability of object center points --- to detect objects and roughly estimate their locations.
Previous work has shown that detecting objects as center points is simple and efficient~\cite{zhou2019objects}.
Moreover, 2D center points can be triangulated to 3D, initializing object poses and enabling geometric reasoning.
For example, center predictions located at impossible locations can be discarded.
When predicting center points, there is a trade-off between spatial precision and generalization.
Precise center locations can differ subtly between similar objects.
This makes it difficult for a learning system to predict precise centers for unseen objects.
As we want our system to generalize to unseen categories, we relax spatial precision requirements, by training our model to predict gaussian blobs at center locations. We use a single center heatmap for all object categories.

\begin{figure}[htb!]
    \centering
    \begin{subfigure}[b]{\linewidth}
        \centering
        \includegraphics[width=\linewidth,keepaspectratio]{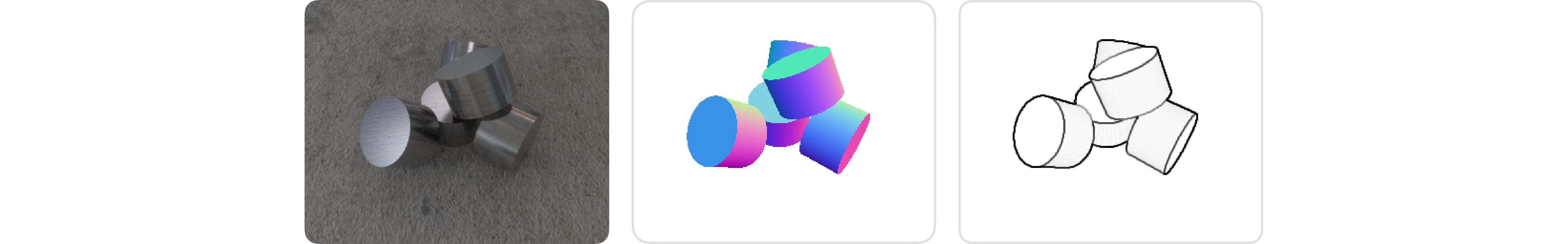}
        \caption{}
        \label{fig:curvatures_1}
    \end{subfigure}
    \begin{subfigure}[b]{\linewidth}
        \centering
        \includegraphics[width=\linewidth,keepaspectratio]{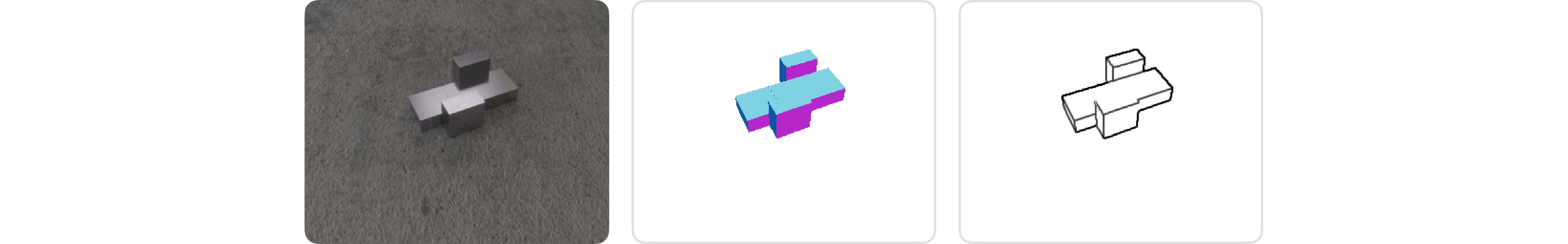}
        \caption{}
        \label{fig:curvatures_2}
    \end{subfigure}
    \caption{
        (a) Curvatures are calculated by rendering normals in view space, approximating per-pixel gradients using the Prewitt operator, and calculating the 2-norm.
        (b) Visually similar areas, like overlapping parallel planes, exhibit no curvature values.
    }
\end{figure}

\paragraph{Curvature}
\label{curvature}
We use curvature heatmaps to highlight local geometry and enable comparison between images and renders.
Representations based on 3D geometry are robust to changes in lighting, texture, and background and can be created from textureless CAD files.
Representations based on global geometry~\cite{wang2019normalized} or category-level semantics~\cite{florence2018dense,peng2019pvnet} do not generalize to unseen object types.
Therefore, we focus on local geometry.
Previous work has shown that geometric edges can be used for accurately estimating 6D poses~\cite{kaskman20206,jensen2022joint,daems2016validation}.
We base our representation on view-space curvature.
To obtain these view-space curvatures, we first render normals in view-space.
Next, we approximate the gradients for each pixel using the Prewitt operator~\cite{prewitt1970object}.
Finally, we calculate the 2-norm of these gradients to obtain a per-pixel curvature value.
Areas with high curvatures correspond to geometric edges or object boundaries and are visually distinct.
Similarly, visually similar areas, like overlapping parallel planes, contain no curvatures values, as shown in figure~\ref{fig:curvatures_2}.

\begin{figure}[htb!]
    \centering
    \includegraphics[width=\linewidth]{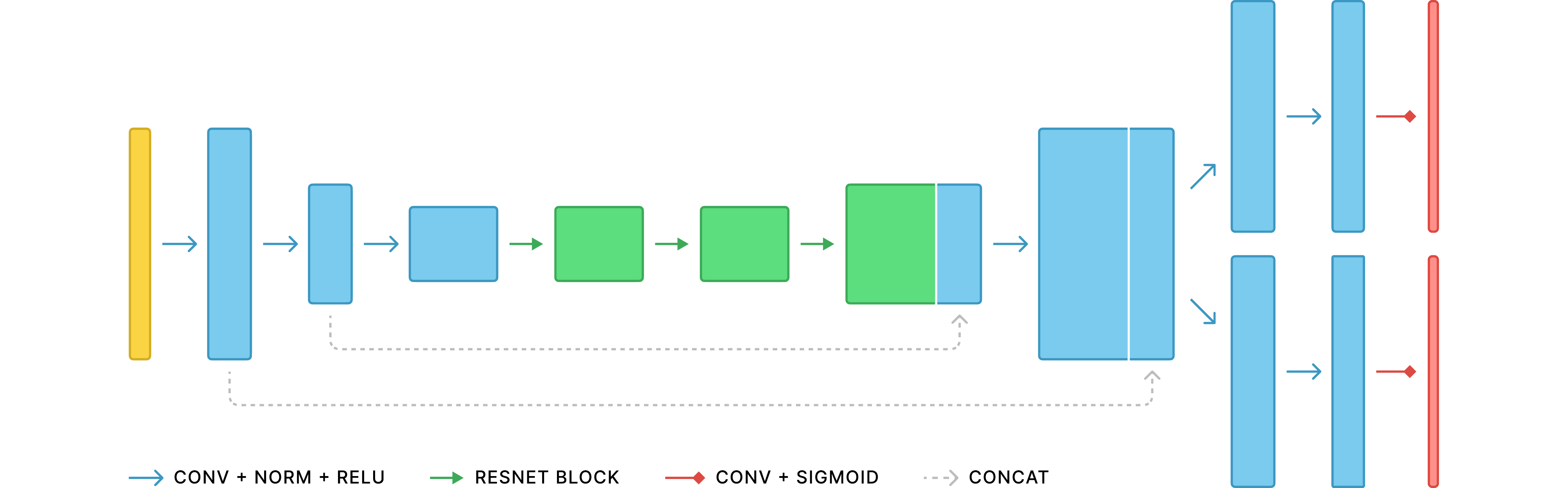}
    \caption{
        The architecture of our fully convolutional network is based on U-net.
        Input images are first processed by a shared backbone.
        Afterwards, separate heads output center and curvature heatmaps.
        \label{fig:model}
    }
\end{figure}

\paragraph{Model and training}
We use a fully convolutional network to predict center and curvature heatmaps.
The same weights are applied to images from different viewpoints.
The architecture is based on U-net~\cite{ronneberger2015u} and shown in figure~\ref{fig:model}.
A shared backbone outputs feature maps with a spatial resolution equal to the input images.
Separate heads are used for predicting center and curvature heatmaps.
Ground-truth center heatmaps are created by projecting 3D object centers to each image and splatting the resulting points using a Gaussian kernel, with standard deviation adapted by object size and distance.
Curvature heatmaps are created as explained in Section~\ref{curvature}.
Binary cross entropy loss is used for both outputs. More details are provided in the appendix.

\subsection{From center heatmaps to 3D centers}

\begin{figure}[htb!]
    \centering
    \includegraphics[width=\linewidth,keepaspectratio]{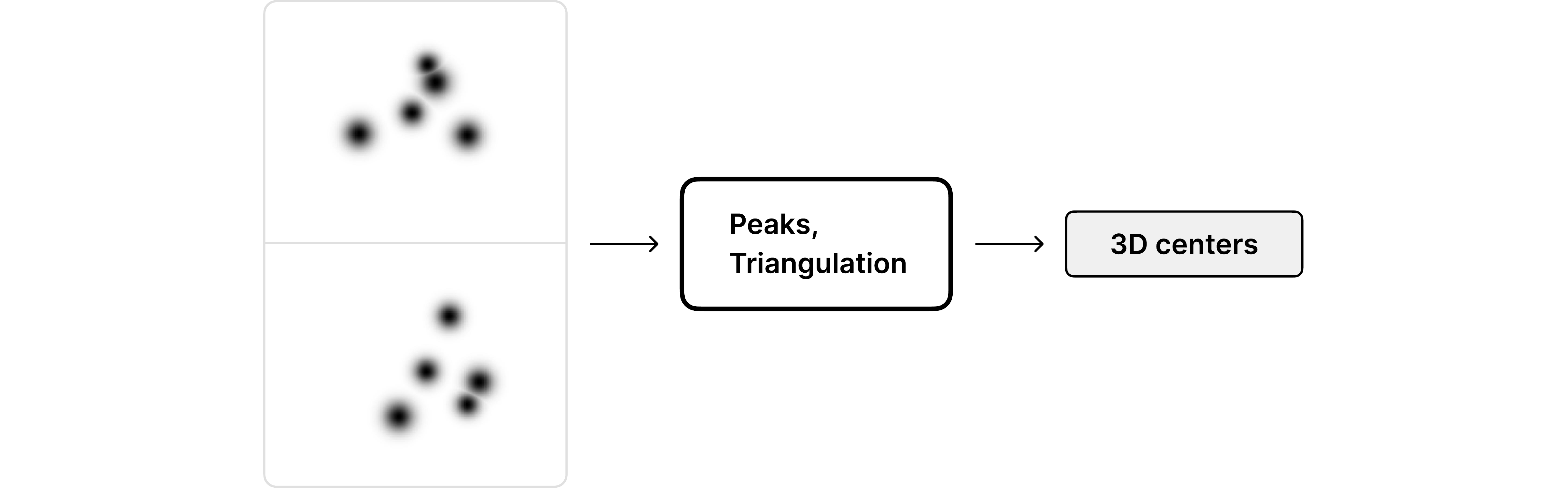}
    \caption{
        Step 2: multi-view 2D center heatmaps are converted to 3D center points.\label{fig:step_2}
    }
\end{figure}

This step converts multi-view center heatmaps to 3D center points.
First, local maxima are found in every heatmap using a peak local max filter~\cite{scikit-image}.
Each of these 2D maxima represents a 3D ray, defined by the respective camera intrinsics and extrinsics.
Next, for each pair of 3D rays, the shortest mutual distance and midpoint are calculated~\cite{Szeliski}.
When this distance is below a threshold $d_t$, the midpoint is added to the set of candidates.
Within this set, points that are closer to each other than a distance $d_c$ are merged.
Finally, the remaining points are refined by maximizing their reprojection score across views, using Scipy's Nelder-Mead optimizer~\cite{2020SciPy-NMeth}.
This leads to a set of 3D points, each with per-view heatmap scores.
If information about the number of objects in the scene is available, this set is further pruned.
3D centers are sorted by their aggregated heatmap score (accumulated for all views) and removed if they are closer than a distance $d_o$ to higher-scoring points.

\subsection{6D pose estimation}

\begin{figure}[htb!]
    \centering
    \includegraphics[width=\linewidth,keepaspectratio]{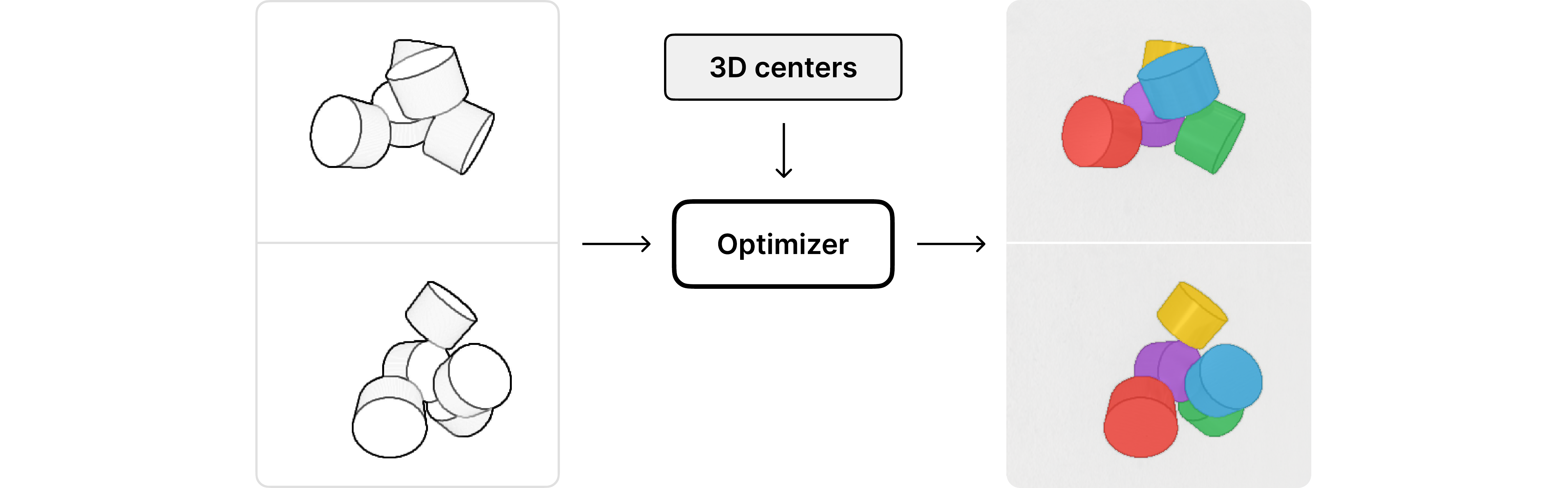}
    \caption{
        Step 3: Multi-view curvature heatmaps and 3D object centers are used to find 6D object poses.
        \label{fig:step_3}
    }
\end{figure}

This step optimizes the 6D poses for all detected objects, using a render-and-compare approach based on curvatures.
Object CAD models, camera intrinsics, and extrinsics are available. Consequently, curvature maps can be rendered for each set of 6D object pose candidates.
We define a cost function that compares such curvature renders to the predicted curvature heatmaps.
This cost function is used for optimizing a set of object poses, initialized by the previously detected 3D centers.

\paragraph{Cost function}
Predicted curvature heatmaps are converted to binary images with threshold $t_b$.
Next, for each binary image, a distance map is created where each pixel contains the distance to the closest non-zero (true) pixel, using scikit-image's distance transform~\cite{scikit-image}.
The resulting distance maps have to be calculated only once and can be reused throughout optimization.

\begin{figure}[htb!]
    \centering
    \includegraphics[width=\linewidth,keepaspectratio]{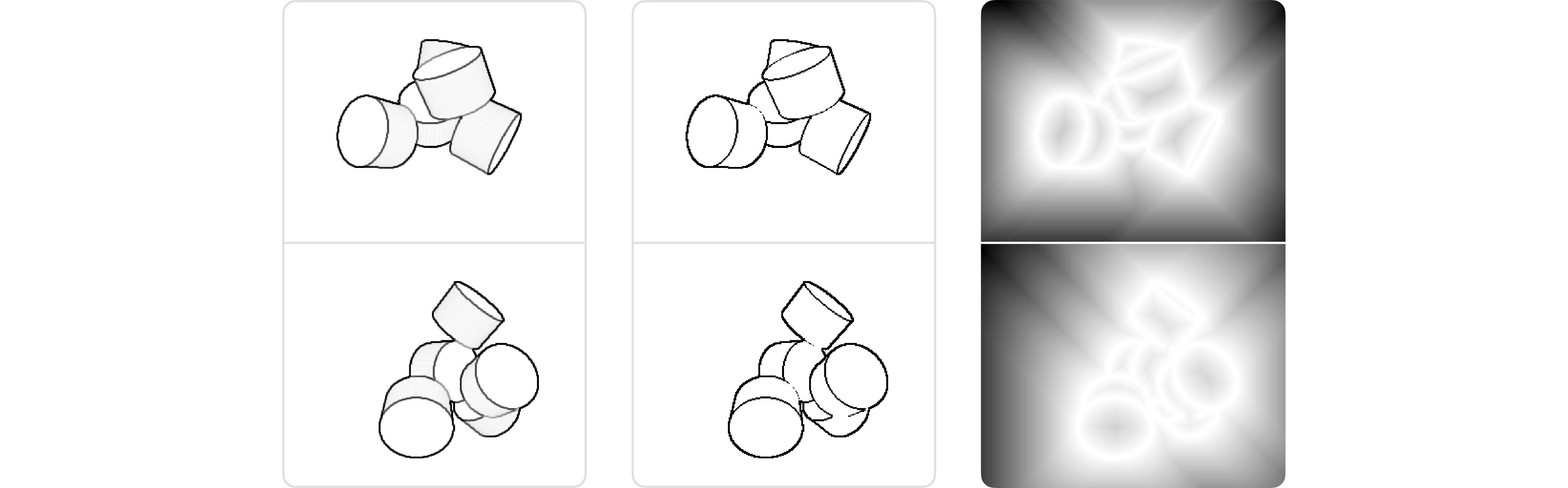}
    \caption{
        Target curvature heatmaps (left) are converted to binary images (center).
        Next, distance maps (right) are calculated, where each pixel contains the distance to the closest non-zero (true) pixel.
        \label{fig:distance_transforms}
    }
\end{figure}

Using these distance maps, curvature renders can be efficiently compared to the target heatmaps.
Pixel-wise multiplication of a render to a distance map returns an image where each pixel contains the distance to the closest true curvature pixel, weighted by its curvature value.
As a result, regions with high curvature weigh more, and regions with zero curvature do not contribute.
The final cost value is obtained by summing the resulting image, and dividing by the sum of the rendered curvature map.
This is done for each viewpoint.
The resulting costs are weighed by view-specific weights $w_v$ and summed, resulting in a final scalar cost value.

\begin{figure}[htb!]
    \centering
    \includegraphics[width=\linewidth,keepaspectratio]{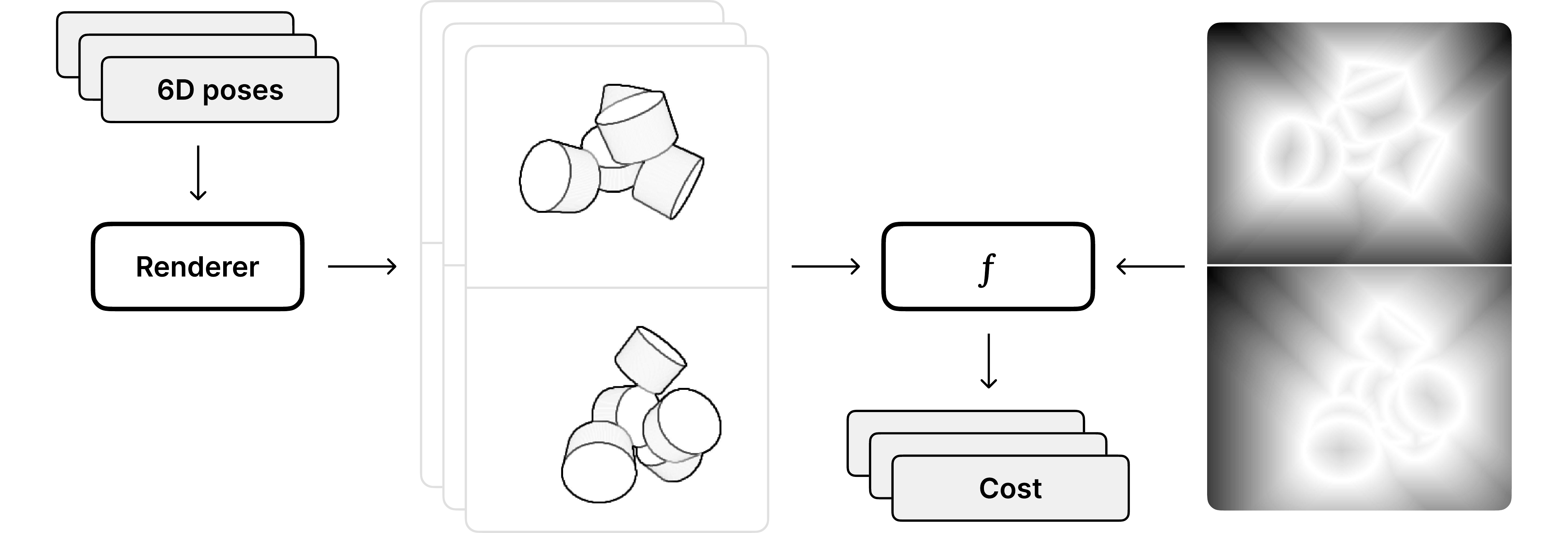}
    \caption{
        Overview of the cost function.
        First, curvature maps are rendered for each set of 6D object pose estimates.
        These curvature maps are compared to a pre-computed distance map based on the target curvatures.
        This results in a cost value for each sample, calculated in parallel.
        \label{fig:cost}
    }
\end{figure}

Figure~\ref{fig:cost} shows an overview of the cost function.
We implement this function ––- including curvature rendering ––- on GPU.
Our implementation runs at 2,000 calls per second on an NVIDIA RTX3090 Ti for six $256\times320$ images per call.

\paragraph{Optimization}
We sequentially optimize 6D object poses by evaluating pose candidates anchored by the detected 3D centers.
As scenes can consist of densely stacked objects with many occlusions, we optimize objects sequentially.
We argue highly visible objects are easier to optimize and --- crucially --- should be taken into account when optimizing objects they occlude.
For each 3D center, we use per-view center heatmap scores as a proxy for visibility.
We optimize objects in order of decreasing visibility, and weigh the contribution of each view to the cost according to these scores.
When estimating an object pose, we first evaluate a set of 2,000 pose candidates with random rotations and translations normally distributed around the 3D center.
Afterwards, the best candidates are further optimized using a bounded Nelder-Mead optimizer~\cite{2020SciPy-NMeth}.

\section{Experiments}

We evaluate our method on the 6D localization task as defined in the BOP challenge~\cite{hodavn2020bop} on DIMO~\cite{deroovere2022dataset} and T-LESS~\cite{hodan2017t}, two industry-relevant datasets.
On DIMO, we show our method significantly outperforms PVNet~\cite{peng2019pvnet}, a strong single-view baseline.
On T-LESS, CenDerNet improves upon the 2020 ECCV results of CosyPose~\cite{labbe2020cosypose}, the leading multi-view method.

\subsection{DIMO}

\paragraph{Dataset}
The dataset of industrial metal objects (DIMO)~\cite{deroovere2022dataset} reflects real-world industrial conditions.
Scenes consist of symmetric, textureless and highly reflective metal objects, stacked in dense compositions.
\paragraph{Baseline}
We compare our method to PVNet with normalized pose rotations.
PVNet~\cite{peng2019pvnet} is based on estimating 2D keypoints followed by perspective n-point optimization.
This method is robuust to occlusion and truncation, but suffers from pose ambiguities caused by object symmetries.
This is amended by mapping rotations to canonical rotations before training \cite{pitteri2019object}.
\paragraph{Experiments}
We use a subset of the dataset, focusing on the most reflective parts, and scenes with objects of the same category.
For training, only synthetic images are used. We report results on both synthetic and real-world test data.
When predicting center and curvature heatmaps, the same output channel is used for all object categories.
Details on data preprocessing and model training can be found in the appendix.

\paragraph{Evaluation}
We report average recall (AR) for two different error functions:
\begin{itemize}
    \item MSPD, the maximum symmetry-aware projection distance, as it is relevant for evaluating RGB-only methods.
    \item MSSD, the maximum symmetry-aware surface distance, as it is relevant for robotic manipulation.
\end{itemize}
Strict thresholds of correctness $\theta_e$ are chosen, as manufacturing use-cases require high-precision poses.
$\theta_{\text{MSSD}}$ is reported for 5\% of the object diameter, $\theta_{\text{MSPD}}$ for $5\text{px}$.
Please see the appendix for more details on the metrics used.

Our system predicts poses in world frame, whereas PVNet --- a single-view method --- predicts per-image poses that are relative to the camera.
For better comparison, we transform our estimated poses to camera frames and compute per-image metrics.

\begin{center}
    \begin{table}
        \centering
        \caption{
            \label{tab:dimo-sim}
            Results on DIMO for synthetic test images.
            CenDerNet significantly outperforms the PVNet baseline.
            The difference between MSSD and MSDP shows the importance of multi-view images for robotics applications where spatial precision is important.
        }
        \begin{tabular}{c | c c}
                                      & $\text{AR}_{\text{MSPD}}<5\text{px}$ & $\text{AR}_{\text{MSSD}}<5\%$ \\
            \hline
            PVNet                     & 0.577                                & 0.079                         \\
            \textbf{CenDerNet} (ours) & \textbf{0.721}                       & \textbf{0.639}                \\
        \end{tabular}
    \end{table}
\end{center}

\begin{center}
    \begin{table}
        \centering
        \caption{
            \label{tab:dimo-real}
            Results on DIMO for real-world test images.
            Both methods are affected by the sim-to-real gap.
            While the results of PVNet plummet, CenDerNet is more robuust.
        }
        \begin{tabular}{c | c c}
                                      & $\text{AR}_{\text{MSPD}}<5\text{px}$ & $\text{AR}_{\text{MSSD}}<5\%$ \\
            \hline
            PVNet                     & 0.016                                & 0.000                         \\
            \textbf{CenDerNet} (ours) & \textbf{0.516}                       & \textbf{0.403}                \\
        \end{tabular}
    \end{table}
\end{center}

\begin{figure}[htb!]
    \centering
    \includegraphics[width=\linewidth,keepaspectratio]{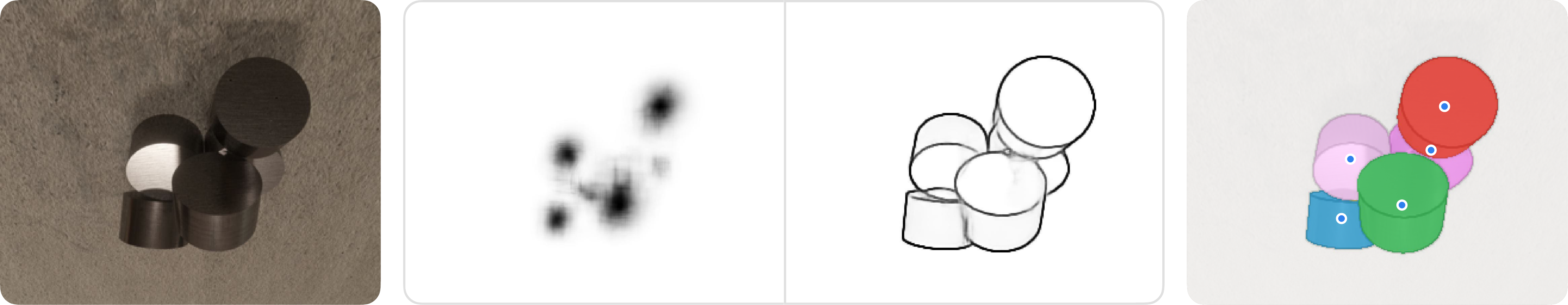}
    \caption{
        Results on synthetic test image of the DIMO dataset.
        Despite severe reflections and shadows, our system recovers qualitative center and curvature representations and precise 6D poses.
        \label{fig:dimo_sim_good}
    }
\end{figure}

\begin{figure}[htb!]
    \centering
    \includegraphics[width=\linewidth,keepaspectratio]{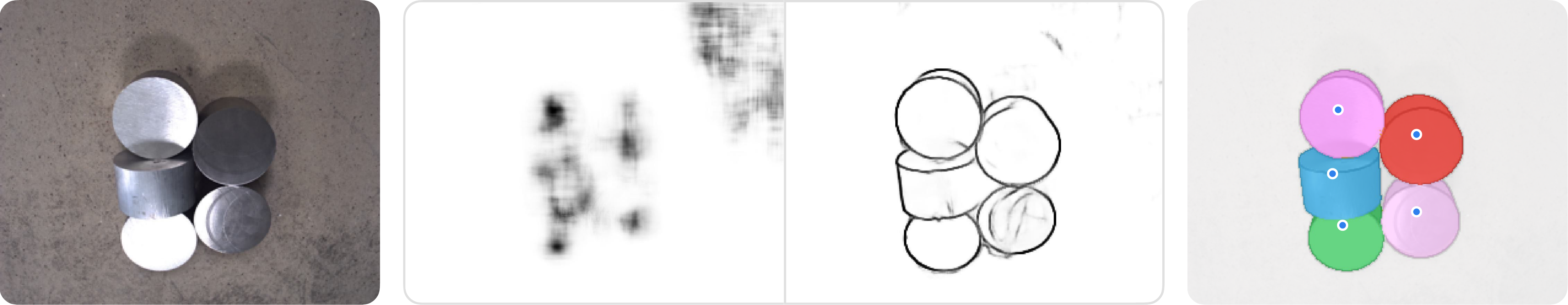}
    \caption{
        Results on real-world test image of the DIMO dataset.
        The lower quality of the center and curvature heatmaps is due to the sim-to-real gap.
        Nevertheless, the estimated 6D poses are accurate.
        \label{fig:dimo_real_good}
    }
\end{figure}

\begin{figure}[htb!]
    \centering
    \includegraphics[width=\linewidth,keepaspectratio]{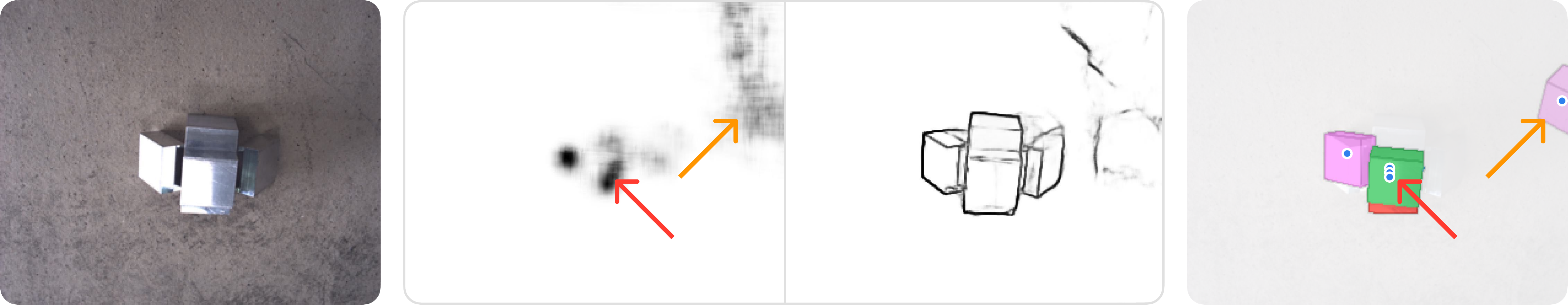}
    \caption{
        Failure case on a real-world test image of the DIMO dataset.
        The unexpected output in the center heatmap (orange arrow) could be due to the sim-to-real gap.
        Objects cannot be differentiated when object centers align (red arrow).
        \label{fig:dimo_real_fail}
    }
\end{figure}

\paragraph{Results}
As shown in Table~\ref{tab:dimo-sim} and Table~\ref{tab:dimo-real}, CenDerNet significantly outperforms the PVNet baseline.
Despite rotation normalizations, PVNet still suffers from pose ambiguities, struggling to predict high-precision poses.

The large gap in MSPD en MSSD scores for PVNet shows the importance of multi-view images for robotics applications.
While PVNet's MSPD score (in image space) is reasonable, its MSSD score (in world/robot space) is subpar.

Reflections and shadows significantly affect many test images, leading to challenging conditions, as shown in Figure~\ref{fig:dimo_sim_good} and Figure~\ref{fig:dimo_real_good}.
Nevertheless, center and curvature predictions are often sufficient to find accurate object poses.
The difference in results between Table~\ref{tab:dimo-sim} and Table~\ref{tab:dimo-real} shows the effect of the sim-to-real gap.
Here, many of the keypoints predicted by PVNet are too far off to meet the strict accuracy requirements, causing results to plummet.
CenDerNet is more robuust.

An important advantage of our method is the interpretability of the predicted center and curvature representations.
The top-right area of the center heatmap in Figure~\ref{fig:dimo_real_good} and Figure~\ref{fig:dimo_real_fail} could be due to the sim-to-real gap.
Adding more background variations while training could improve the robuustness of the model.

Figure~\ref{fig:dimo_real_fail} shows a failure case.
When object centers align, the system fails to differentiate between them.

\subsection{T-LESS}

\paragraph{Dataset}
The T-LESS dataset provides multi-view images of non-reflective, colorless industrial objects.
As part of the BOP benchmark, it allows easy comparison to state-of-the-art methods.
For the 2020 BOP challenge \cite{hodavn2019photorealistic}, photorealistic synthetic training images (PBR) were provided.

\paragraph{Baseline}
We compare our method to CosyPose \cite{labbe2020cosypose}, the multi-view method that won the 2020 BOP challenge.
We use the ECCV 2020 BOP evaluation results for 8 views made available by the authors.

\paragraph{Experiments}
We use only PBR images for training.
As each scene contains a mix of object types, we extend our model to predict a separate center heatmap for each object category.
This reduces the generality of our method, but allows us to select the correct CAD model when optimizing poses.
As training images contain distractor objects, we take object visibility into account when generating ground-truth center and curvature heatmaps.
Details on data preprocessing and model training can be found in the appendix.

\paragraph{Evaluation}
In addition to reporting $\text{AR}_{\text{MSPD}}<5\text{px}$ and $\text{AR}_{\text{MSSD}}<5\%$, we report the default BOP~\cite{hodavn2020bop} evaluation metrics (VSD, MSSD and MSPD).
We transform poses to camera-frames and calculate metrics for all images.

\begin{center}
    \begin{table}
        \centering
        \caption{
            \label{tab:tless-metrics}
            Results on TLESS.
            CenDerNet outperforms CosyPose on both metrics.
            The stark difference in MSSD score is relevant for high-precision robotics applications.
        }
        \begin{tabular}{c | c c}
                                      & $\text{AR}_{\text{MSPD}}<5\text{px}$ & $\text{AR}_{\text{MSSD}}<5\%$ \\
            \hline
            CosyPose (ECCV 2020)      & 0.499                                & 0.250                         \\
            \textbf{CenDerNet} (ours) & \textbf{0.543}                       & \textbf{0.544}                \\
        \end{tabular}
    \end{table}
\end{center}

\begin{center}
    \begin{table}
        \centering
        \caption{
            \label{tab:cosypose}
            Results on TLESS for the default BOP evaluation metrics.
            CenDerNet outperforms the provided CosyPose results.
            However, the scores reported on the BOP leaderboard are still significantly higher.
        }
        \begin{tabular}{c | c c c c}
                                       & $\text{AR}$    & $\text{AR}_{\text{MSPD}}$ & $\text{AR}_{\text{MSSD}}$ & $\text{AR}_{\text{VSD}}$ \\
            \hline
            CosyPose (ECCV 2020)       & 0.617          & 0.686                     & 0.610                     & 0.557                    \\
            \textbf{CenDerNet} (ours)  & 0.713          & 0.715                     & 0.717                     & 0.707                    \\
            CosyPose (BOP leaderboard) & \textbf{0.839} & \textbf{0.907}            & \textbf{0.836}            & \textbf{0.773}           \\
        \end{tabular}
    \end{table}
\end{center}

\begin{figure}[htb!]
    \centering
    \includegraphics[width=\linewidth,keepaspectratio]{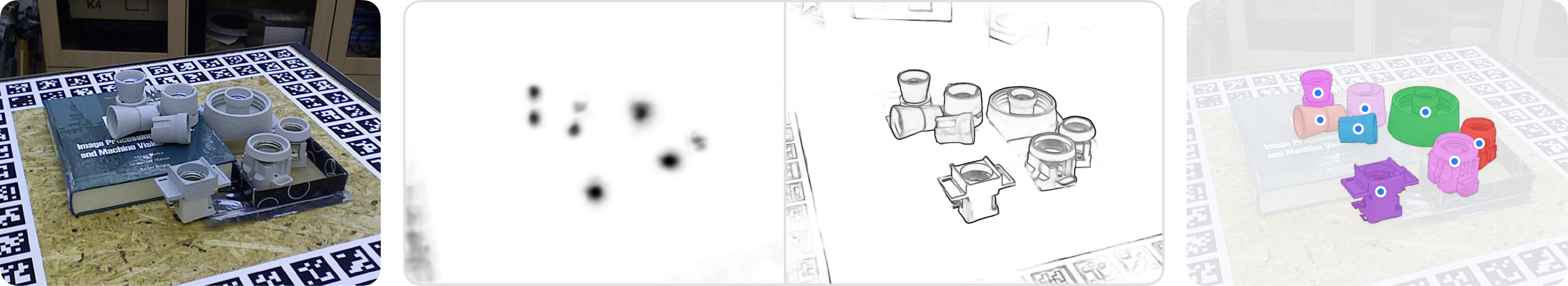}
    \caption{
        Results on T-LESS.
        Our model is able to eliminate irrelevant scene elements, like backgrounds and distractor objects.
        \label{fig:tless_good}
    }
\end{figure}

\begin{figure}[htb!]
    \centering
    \includegraphics[width=\linewidth,keepaspectratio]{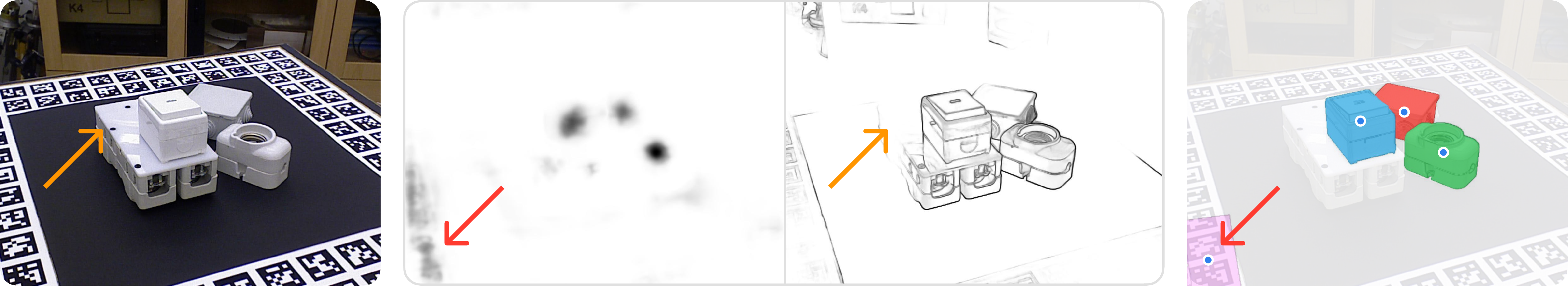}
    \caption{
        Failure case on T-LESS.
        The model is confused about the charuco tags around the scene, leading to an incorrect 3D center.
        The pose optimization process cannot recover from this error.
        \label{fig:tless_fail}
    }
\end{figure}

\paragraph{Results}

As shown in Table~\ref{tab:tless-metrics}, our method outperforms CosyPose in high-precision scenario's.
Table~\ref{tab:cosypose} shows we also outperform the provided CosyPose predictions on the default BOP metrics.
However, the CosyPose-variant that is currently at the top of the BOP leaderboard still performs significantly better.

Figure~\ref{fig:tless_good} shows predicted heatmaps, curvatures object centers and poses.
In many situations, our method is capable of eliminating irrelevant backgrounds and objects, despite being trained only on synthetic data.
There is, however, still a sim-to-real gap, as shown in Figure~\ref{fig:tless_fail}.
Again, the predicted representations provide hints on how to improve the system.
In this case, the model is confused by charuco tags on the edge of the table.

\section{Conclusions}
We present CenDerNet, a system for multi-view 6D pose estimation based on center and curvature representations.
Our system is conceptually simple and therefore easy-to-use.
First, a single neural network converts images into interpretable representations.
Next, a render-and-compare approach, with GPU-optimized cost function, allows for jointly optimizing object poses across multiple viewpoints, thereby naturally handling occlusions and object symmetries.
We demonstrate our system on two challenging, industry-relevant datasets and show it outperforms PVNet, a strong single-view baseline, and CosyPose, the leading multi-view approach.
In future work, we plan to explore ways to further decrease the processing time of our method.
In addition, we will investigate improvements to the robustness of our pipeline.
For example, fusing information from multiple viewpoints could improve center and curvature predictions.
We also look forward to evaluating our method in a real-world setup.

\section*{Acknowledgement}
The authors wish to thank everybody who contributed to this work.
The authors thank RoboJob (\url{https://www.robojob.eu}) for their support and the members of the \textit{Keypoints Gang} for many insightfull discussions.
This research was supported by VLAIO Baekeland Mandate HBC.2019.2162 and by Flanders Make (\url{http://www.flandersmake.be}) through the PILS SBO project (Project Inspection with Little Supervision)
and the CADAIVISION SBO project.
Furthermore this research received funding from the Flemish Government under the ``Onderzoeksprogramma Artificiële Intelligentie (AI) Vlaanderen'' programme.

\clearpage
%
%
\bibliographystyle{splncs04}
\bibliography{references}

\clearpage

\appendix

\section{Pose error functions}

We use MSDP and MSSD as defined by the BOP challenge \cite{hodavn2020bop} to estimate pose errors, with implementations provided by the BOP Toolkit~\cite{hodan2020boptoolkit}.

\paragraph{Symbols}
The following symbols are used:
\begin{itemize}
    \item $\hat{\textbf{P}}$ estimated pose
    \item $\bar{\textbf{P}}$ ground-truth pose pose
    \item $V_M$ set of mesh vertices of the object model
    \item $S_M$ set of symmetry transformation of the object model
    \item $\text{proj}$ result of the 2D projection operation
\end{itemize}

\paragraph{MSPD}
The Maximum Symmetry-aware Projection Distance:
\begin{equation}
    \text{MSPD}\big(\hat{\mathbf{P}}, \bar{\mathbf{P}}, S_M, V_M\big) = \text{min}_{\textbf{S} \in S_M} \text{max}_{\textbf{x}
    \in V_M}
    \big\Vert \text{proj}\big( \hat{\textbf{P}}\textbf{x} \big) - \text{proj}\big(
    \bar{\textbf{P}}\textbf{S}\textbf{x} \big) \big\Vert_2
\end{equation}

\paragraph{MSSD}
The Maximum Symmetry-aware Surface Distance:
\begin{equation}
    \text{MSSD}\big(\hat{\mathbf{P}}, \bar{\mathbf{P}}, S_M, V_M\big) = \text{min}_{\textbf{S} \in S_M} \text{max}_{\textbf{x}
    \in V_M}
    \big\Vert \hat{\textbf{P}}\textbf{x} - \bar{\textbf{P}}\textbf{S}\textbf{x}
    \big\Vert_2
\end{equation}

\section{Technical details}

This section will provide more details about the data processing, model architecture, and hyperparameters used in our experiments.

\begin{figure}[htb!]
    \caption{
        Model architecture.
        The shape of the different feature maps will be reported using the following indications.
        \label{fig:model_labeled}
    }
    \centering
    \includegraphics[width=\linewidth]{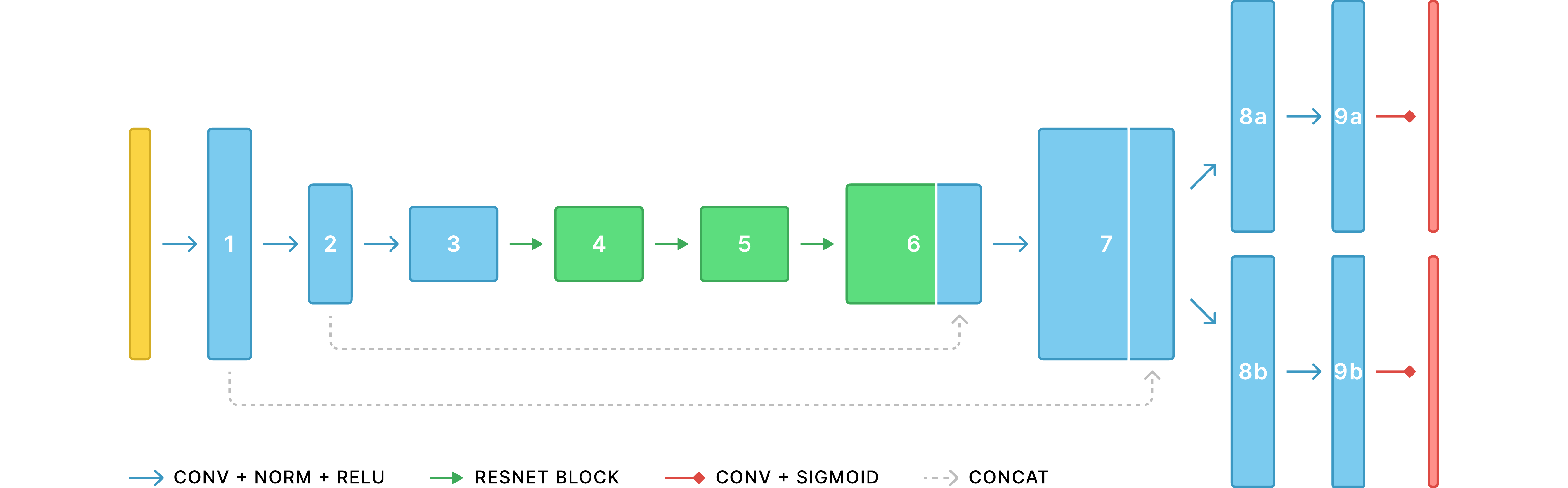}
\end{figure}

\subsection{Data}
For all experiments, the following data augmentations are used.
Note that as the task of the model is to predict center and curvature representations, we can apply augmentations that significantly affect object pose information, such as flipping.
\begin{itemize}
    \item horizontal flip $p=0.5$
    \item vertical flip $p=0.5$
    \item scale $U(1.0,1.25)$
    \item brightness $U(0.7,1.3)$
    \item contrast $U(0.7,1.3)$
    \item hue $U(-0.1,0.1)$
    \item gaussian blur, $\text{kernel}=3$, $\text{sigma}=0.05$, $p=0.5$
    \item gaussian noise $p=0.5$
    \item sharpness, $U(0.5,1.5)$, $p=0.5$
\end{itemize}
When creating ground-truth center heatmaps, we create gaussian blobs with sigma calculated as follows:
\begin{equation}
    \sigma = s_e \times \frac{\text{mean object bbox extents} \times 100}{\text{distance center to camera}}
\end{equation}

\subsection{DIMO}
\paragraph{Data}
We use the small version the DIMO dataset, with $256\times320$ images.
A subset of the data is used by selecting scenes using the cardboard carrier (\texttt{carrier id} 2), and containing only metal objects of the same type (\texttt{composition id} 1, 2, and 4).
For each scene, seven viewpoints are selected (\texttt{viewpoint id} 0, 2, 4, 6, 8, 10, 12).
For training, we use synthetic images (\texttt{sim\_jaigo}) with random poses and random lighting (\texttt{scene id} with suffix 16-72).
For testing, we report results for both synthetic (\texttt{sim\_jaigo}) and real (\texttt{real\_jaigo}) images (\texttt{scene id} with suffix 00).
When creating ground-truth center heatmaps, we use $s_e=0.5$.

\paragraph{Model}
Our model architecture leads to the following feature maps:

\begin{center}
    \begin{table}
        \centering
        \begin{tabular}{r | c c c c c c c c c c c}
                     & input & 1   & 2   & 3  & 4  & 5       & 6       & 7   & 8                                            & 9                                            & output                                     \\
            \hline
            height   & 256   & 256 & 128 & 64 & 64 & 64      & 128     & 256 & 256                                          & 256                                          & 256                                        \\
            width    & 320   & 320 & 160 & 80 & 80 & 80      & 160     & 320 & 320                                          & 320                                          & 320                                        \\
            channels & 3     & 32  & 32  & 64 & 64 & 64 + 32 & 64 + 32 & 64  & \begin{tabular}{@{}c@{}}32\\ 32\end{tabular} & \begin{tabular}{@{}c@{}}32\\ 32\end{tabular} & \begin{tabular}{@{}c@{}}1\\ 1\end{tabular}
        \end{tabular}
        \caption{
            Feature map sizes for the model used for DIMO.
            \label{tab:dimo_model}
        }
    \end{table}
\end{center}

\paragraph{Hyperparameters}
We train for 200 epochs with a batch size of 8 and learning rate of 3e-4. Training takes 8 hours on a NVIDIA RTX 3090 Ti.
When converting center heatmaps to 3D centers, we use $d_t=30$, $d_c=30$ and $d_o=30$.

\subsection{T-LESS}

\paragraph{Data}
We use all PBR images for training. For testing, we randomly select 5 views to predict scene poses.
When creating ground-truth center heatmaps, we use $s_e=1.25$.
Blobs are only added for objects with a visibility score $v > 0.5$.
When creating ground-truth curvature maps, we apply the provided visibility masks to ensure only visible curvatures are present.

\paragraph{Model}
Our model architecture leads to the following feature maps:

\begin{center}
    \begin{table}
        \centering
        \begin{tabular}{r | c c c c c c c c c c c}
                     & input & 1   & 2   & 3   & 4   & 5         & 6        & 7   & 8                                              & 9                                              & output                                      \\
            \hline
            height   & 540   & 180 & 180 & 60  & 60  & 60        & 180      & 540 & 540                                            & 540                                            & 540                                         \\
            width    & 720   & 240 & 240 & 80  & 80  & 80        & 240      & 720 & 720                                            & 720                                            & 720                                         \\
            channels & 3     & 64  & 128 & 256 & 256 & 256 + 128 & 256 + 64 & 256 & \begin{tabular}{@{}c@{}}128\\ 128\end{tabular} & \begin{tabular}{@{}c@{}}128\\ 128\end{tabular} & \begin{tabular}{@{}c@{}}1\\ 30\end{tabular}
        \end{tabular}
        \caption{
            Feature map sizes for the model used for T-LESS. For centers, 30 output channels are used (one for each object category), for curvatures 1 shared output channel is used.
            \label{tab:tless_model}
        }
    \end{table}
\end{center}

\paragraph{Hyperparameters}
We train for 40 epochs with a batch size of 4 and learning rate of 3e-4. Training takes 6 days on an NVIDIA Tesla V100.
When converting center heatmaps to 3D centers, we use $d_t=30$, $d_c=30$ and $d_o=30$.

\paragraph{Other}
When comparing to CosyPose, we use the \texttt{cosypose114533\-eccv2020\_tless\-test\-primesense} results, as provided on the CosyPose GitHub page.
\end{document}